\documentclass[runningheads]{llncs}
\usepackage{graphicx}
\usepackage{tikz}
\usepackage{comment}
\usepackage{amsmath,amssymb} % define this before the line numbering.
\usepackage{color}
\usepackage{multirow}
\usepackage{graphicx}
\usepackage{subfigure}
\usepackage{algorithm}
\usepackage{algorithmic}
\usepackage{hyperref}
\usepackage{caption}
\hypersetup{colorlinks=true,urlcolor=blue}

\def\eg{\emph{e.g}\onedot} 

\def\ie{\emph{i.e}\onedot}

\def\ie{\mbox{\textit{i.e.}, }}
\def\eg{\mbox{\textit{e.g.}, }}

\def\CE{\text{CE}}
\def\KL{\text{KL}}
\def\onehot{\mathrm{CE}}
\def\kd{\mathrm{KD}}
\def\bns{\mathrm{BNS}}

\def\bmu{\mbox{{\boldmath $\mu$}}}

\def\bsigma{\mbox{{\boldmath $\sigma$}}}

\def\mL{{\mathcal L}}

\def\mN{{\mathcal N}}

\DeclareMathAlphabet\mathbfcal{OMS}{cmsy}{b}{n}
\def\0{{\bf 0}}
\usepackage{ dsfont }
\def\1{\mathds{1}}

\def\bx{{\bf x}}

\def\bz{{\bf z}}

\def\mmE{{\mathbb E}}

\def\bx{{\bf x}}

\def\bz{{\bf z}}

\usepackage{ntheorem}

\usepackage{enumitem}

\def\kui{\textcolor{black}}

\def\mytitle{\textcolor{black}{Generative Low-bitwidth Data Free Quantization }}

\def\lhk{\textcolor{black}}
\def\xsk{\textcolor{black}}
\def\jing{\textcolor{black}}

\def\jie{\textcolor{black}}

\def\wrong{\textcolor{black}}
\def\new{\textcolor{black}}
\definecolor{mypink}{cmyk}{0, 0.7808, 0.4429, 0.1412}

\begin{document}
% \linenumbers
\pagestyle{headings}
\mainmatter
\def\ECCVSubNumber{1469}  % Insert your submission number here

\title{Generative Low-bitwidth Data Free Quantization} 

% Replace with your title

\begin{comment}
\titlerunning{ECCV-20 submission ID \ECCVSubNumber} 
\authorrunning{ECCV-20 submission ID \ECCVSubNumber} 
\author{Anonymous ECCV submission}
\institute{Paper ID \ECCVSubNumber}
\end{comment}
%******************

% \begin{comment}
\titlerunning{Generative Low-bitwidth Data Free Quantization} 

\author{Shoukai Xu\inst{1,2}$^*$  \and 
	Haokun Li\inst{1}$^*$  \and 
	Bohan Zhuang\inst{3}\thanks{Authors contributed equally.} \and   \\
	Jing Liu\inst{1} \and
	Jiezhang Cao\inst{1}  \and 
	Chuangrun Liang\inst{1} \and 
	Mingkui Tan\inst{1}\thanks{Corresponding author.}
}

\authorrunning{Shoukai Xu, Haokun Li, Bohan Zhuang, Mingkui Tan, and et al.}
\institute{
South China University of Technology, Guangzhou, China
\\ \email{\{sexsk,selihaokun,secaojiezhang,seliujing,selcr\}@mail.scut.edu.cn} \email{mingkuitan@scut.edu.cn} \\
\and 
PengCheng Laboratory, Shenzhen, China
\and Monash University, Melbourne, Australia \\
\email{bohan.zhuang@monash.edu}
}

% \end{comment}

\makeatletter
\renewcommand*{\@fnsymbol}[1]{\ensuremath{\ifcase#1\or *\or \dagger\or \ddagger\or
		\mathsection\or \mathparagraph\or \|\or **\or \dagger\dagger
		\or \ddagger\ddagger \else\@ctrerr\fi}}
\makeatother

%******************
\maketitle

\begin{abstract}

\kui{Neural network quantization is an effective way to compress deep models and improve \new{their} execution latency and energy efficiency, so that they can be deployed on mobile or embedded devices.}
\xsk{Existing quantization methods require original data for calibration or fine-tuning to get better performance.}
However, in many real-world scenarios, the data may not be available due to confidential or private issues, \new{thereby} making existing quantization methods not applicable. \kui{Moreover, due to the absence of original data, the recently developed generative adversarial networks (GANs) \new{cannot} be applied to generate data.}    
\kui{Although the full-precision model may contain \new{rich} data information, such information alone is hard to exploit for recovering the original data or generating new meaningful data.}  
In this paper, we investigate a simple-yet-effective method called \mytitle\new{(GDFQ)} to remove the data dependence burden. 
Specifically, we propose a knowledge matching generator to produce meaningful fake data by exploiting classification boundary knowledge and distribution information in the pre-trained model.
With the help of generated data, we \new{can} quantize a model by learning knowledge from the pre-trained model.
Extensive experiments on three data sets demonstrate the effectiveness of our method. 
More critically, our method achieves much higher accuracy on 4-bit quantization than the existing data free quantization method.
\new{Code is available at \url{https://github.com/xushoukai/GDFQ}.}


\end{abstract}

\begin{figure}[t]
\centering
\subfigure[\scriptsize{Real data}]{
\includegraphics[width=1.15in]{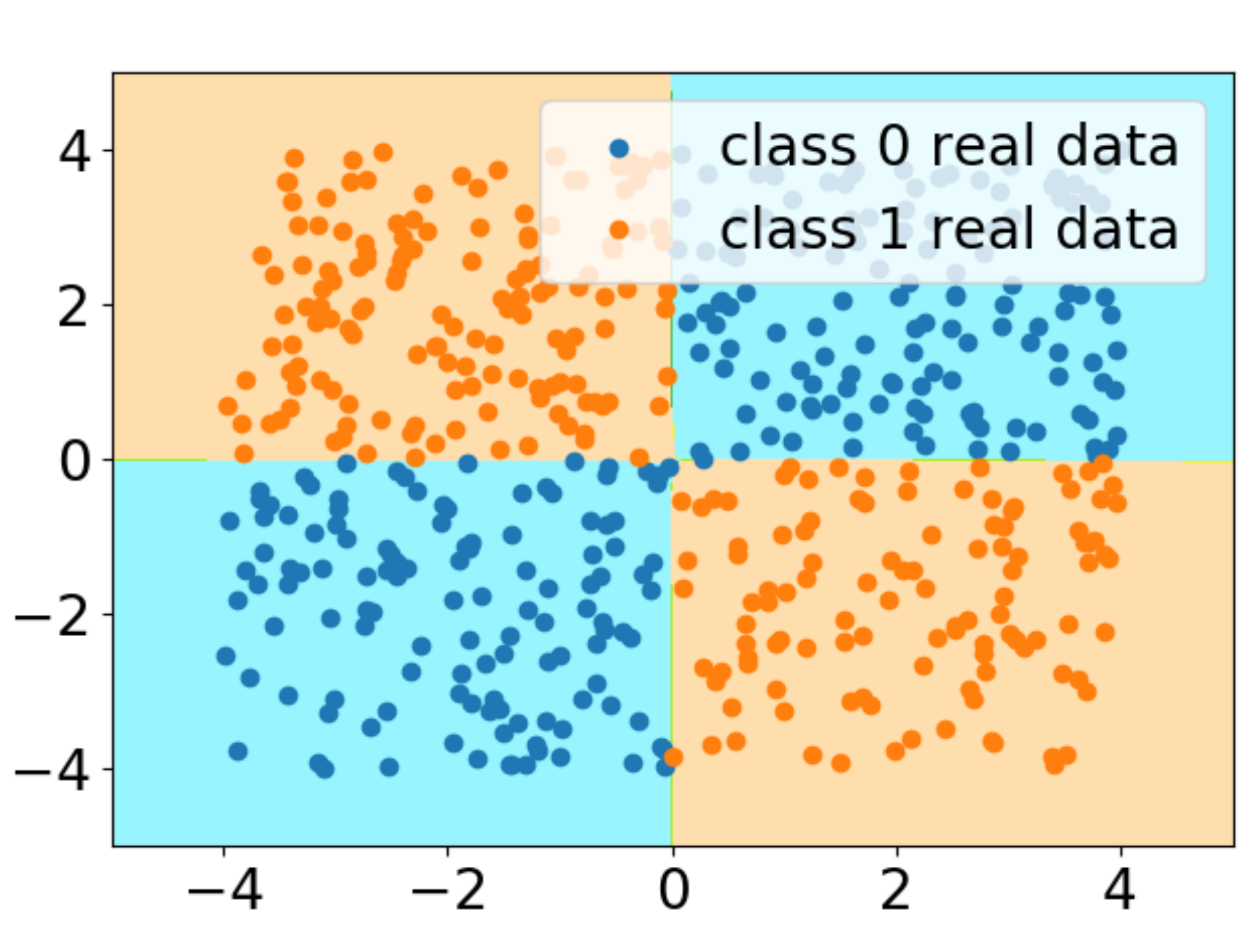}
\hspace{-1.5mm}
}%
\subfigure[\scriptsize{Gaussian input data}]{
\includegraphics[width=1.15in]{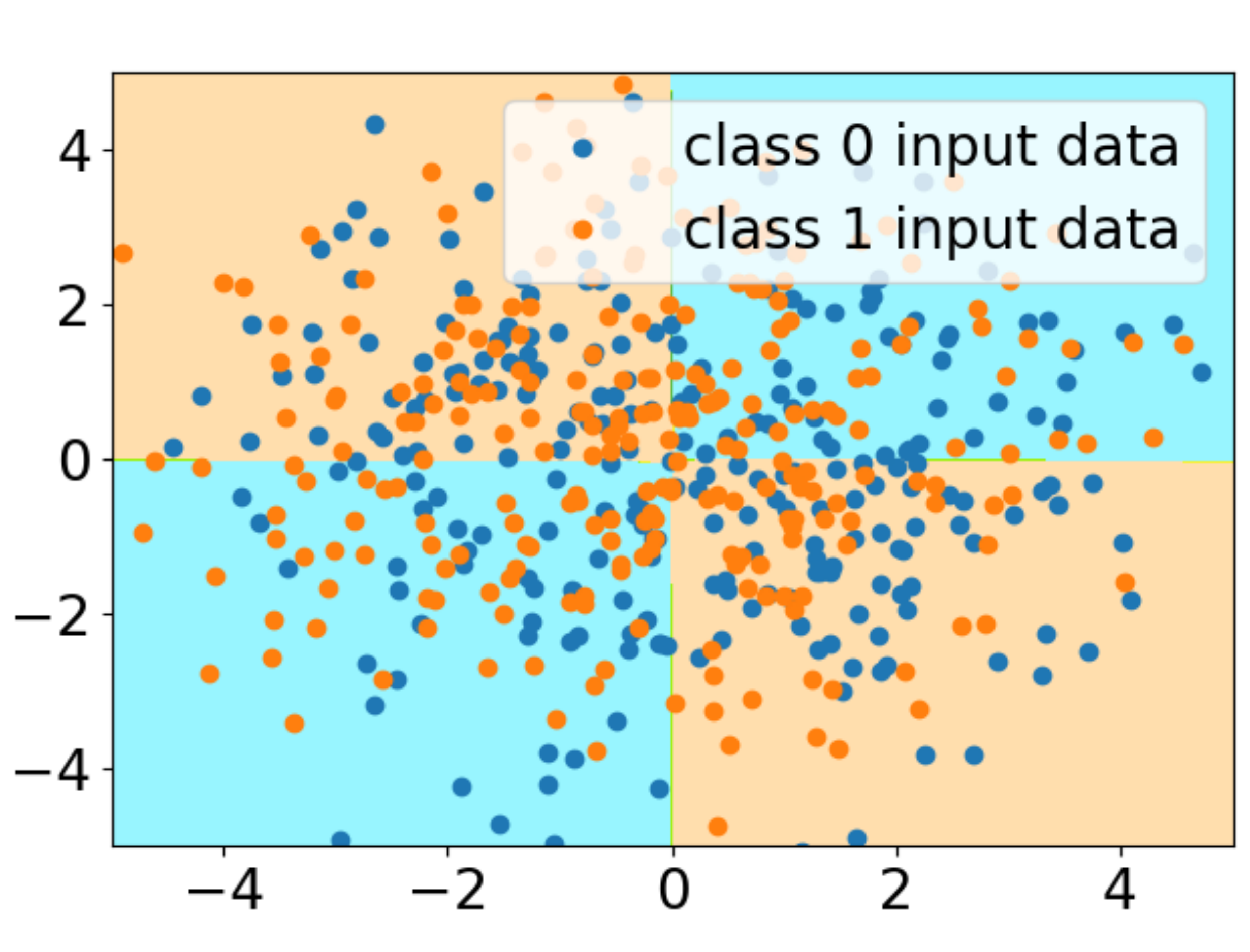}
\hspace{-1.0mm}
}%
\subfigure[\scriptsize{Fake data (ZeroQ~\cite{Cai_2020_CVPR})}]{
\includegraphics[width=1.15in]{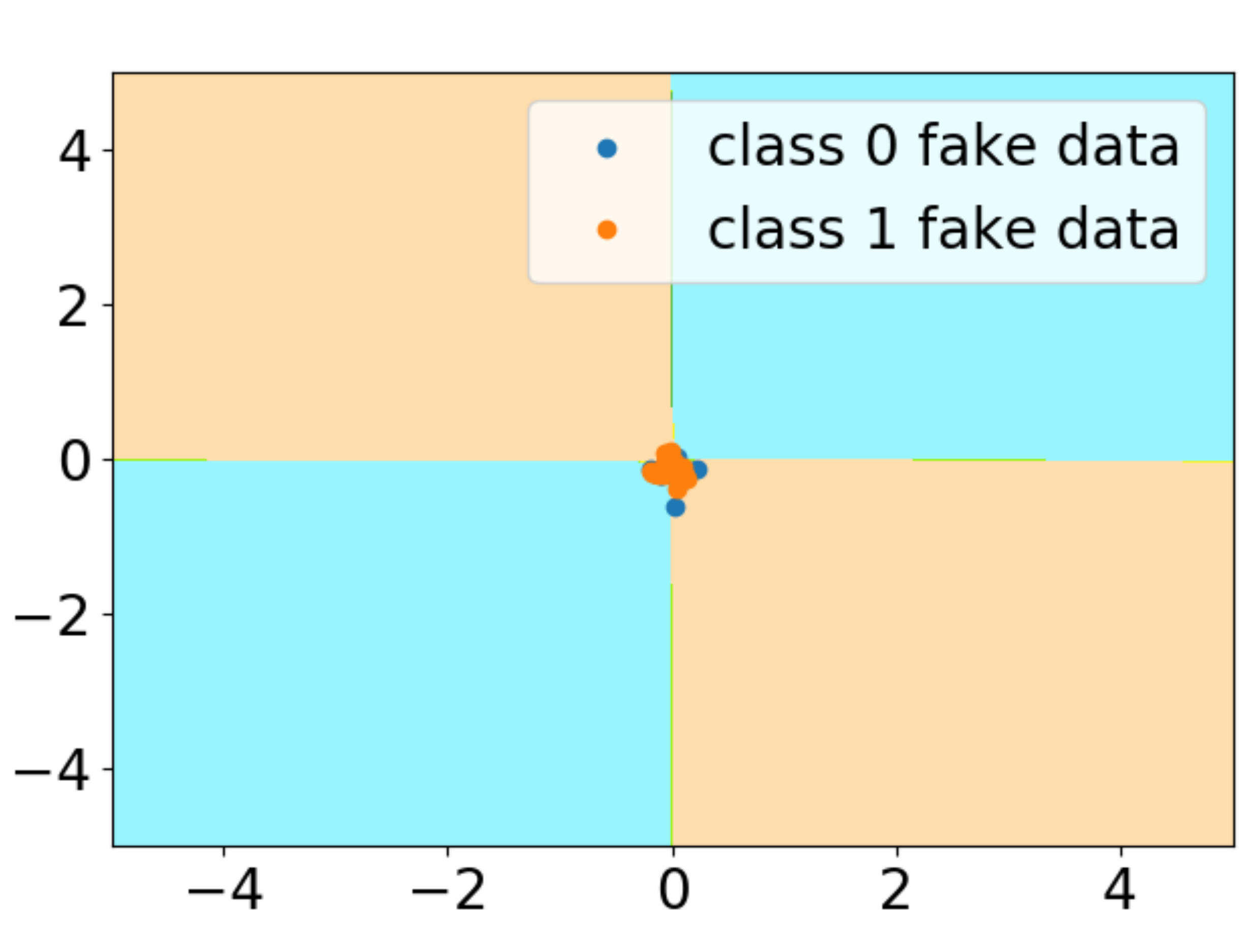}
\hspace{-1.5mm}
}%
\subfigure[\scriptsize{Fake data (Ours)}]{
\includegraphics[width=1.15in]{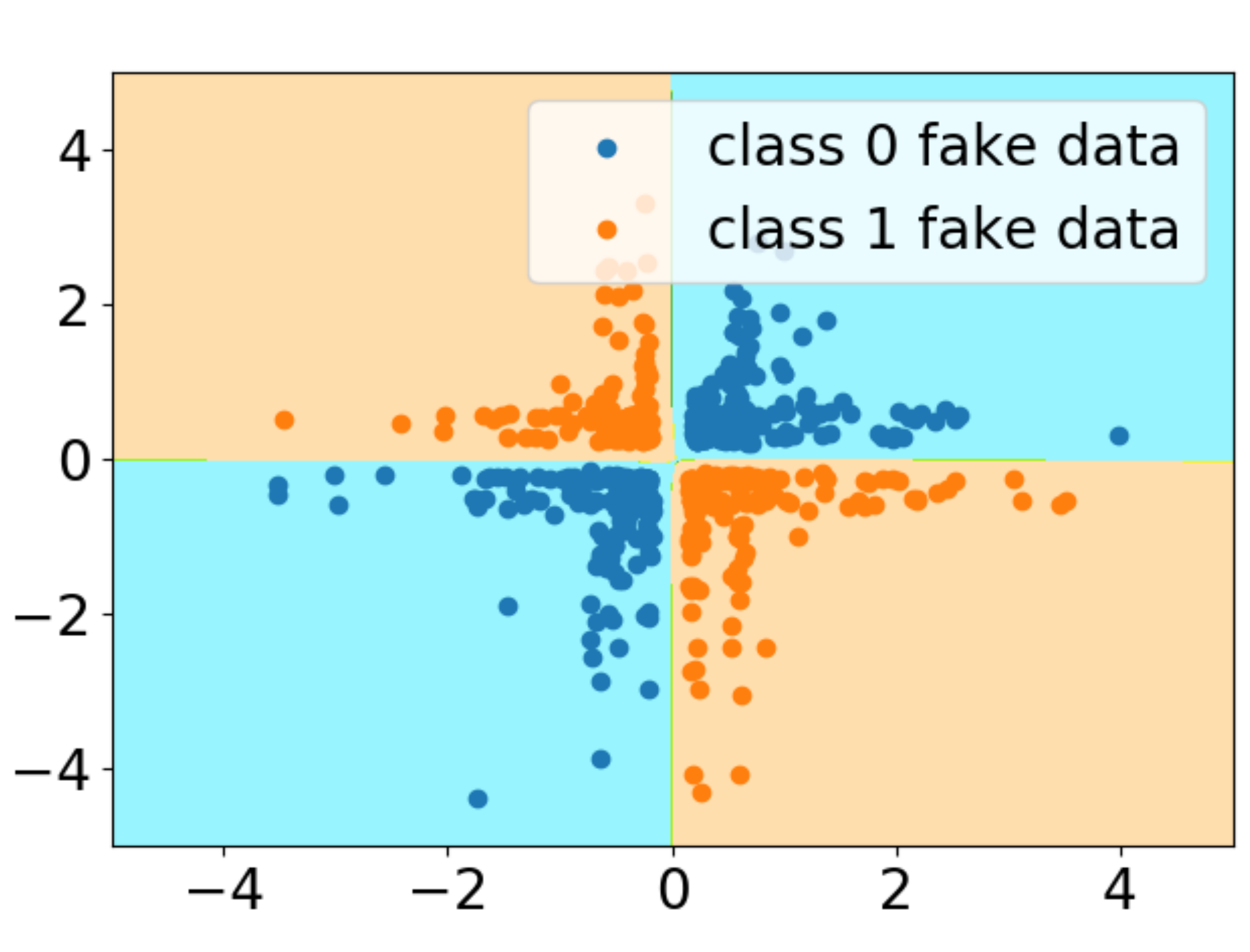}
}%
\centering
\caption{The comparisons \new{of} generated fake data. 
\new{ZeroQ generates fake data by gradient updating; however, since it neglects the inter-class information, it is hard to capture the original data distribution.}
\new{Meanwhile, the} knowledge matching generator produces fake data with label distribution\new{,} and it is more likely to produce data distributed near \new{the} classifier boundaries.}
\label{motivation}
\end{figure}

\section{Introduction}
\label{Introduction}
Deep neural networks (DNNs) have achieved great success in many areas, such as computer vision \cite{he2016deep,Szegedy2016inceptionv3,sandler2018mobilenetv2,guo2019nat,zeng2019graph} and  natural language processing \cite{mikolov2010recurrent,sak2014long,devlin2018bert}.
\kui{However, DNNs often contain a \new{considerable} number of parameters, which makes them hard to be deployed on embedded/edge devices due to unbearable computation \new{costs} for inference. 
Network quantization, which aims to reduce the model size by quantizing floating-point values into low precision(\eg 4-bit), is an effective way to improve the execution latency and energy efficiency.
Existing quantization methods \cite{zhou2016dorefa,choi2018pact,zhuang2018towards,zhao2019improving,lin2019defensive} generally require training data for calibration or fine-tuning. 
Nevertheless, in many real-world applications in medical \cite{zhang2019whole}, finance \cite{zhang2018adaptive} and industrial domains, the training data may not be available due to privacy or confidentiality issues.
Consequently, these methods are no longer applicable due to the absence of training data. Therefore, the data-free quantization is of great practical value.}

\kui{To address this \new{issue}, one possible way is to directly sample random inputs from some distribution and then use \new{these inputs} to quantize the model so that the output distributions of the full-precision model and quantized model are as close as possible.}
Unfortunately, since random inputs contain no semantic information and are far away from the real data distribution, this simple \new{method} suffers from huge performance degradation .
Alternatively, one can use generative adversarial networks (GANs) to produce data.
\new{Due to the absence of training data, GANs cannot be applied in the data-free quantization.}
In practice, the quantization performance highly depends on the quality of the input data.
Therefore, \new{the process of constructing} meaningful data to quantize models is very challenging.

To generate meaningful data, it is important and necessary to exploit data information from a pre-trained model.
\new{A} recent study \cite{zhang2016understanding} \new{revealed} that a well-trained over-parameterized model maintains sufficient information about the entire data set.
Unfortunately, what information exists and how to exploit such information are still unknown.
In the training, a neural network uses batch normalization \cite{ioffe2015batch} to stabilize the data distribution and learns a classification boundary to divide data into different classes (See Fig. \ref{motivation} (a)).
In this sense, some information about the classification boundary and data distribution is hidden in the pre-trained model.
However, these kinds of information are disregarded by existing data-free quantization methods \cite{Cai_2020_CVPR,nagel2019data}\new{,} which only focus on a single sample or network parameters.
For example, ZeroQ \cite{Cai_2020_CVPR} exploits the information of a single sample instead of entire data, \new{causing} the constructed distribution to be far away from the real data distribution (See Fig. \ref{motivation} (c)).
To address these issues, how to construct meaningful data by fully exploiting the classification boundary and distribution information from a pre-trained model remains an open question.

In this paper, we propose a simple-yet-effective method called \mytitle to achieve completely data-free quantization.
\jie{Without original data, we aim to learn a good generator to produce meaningful data by mining the classification boundary knowledge and distribution information in batch normalization statistics (BNS) from the pre-trained full-precision model.}

\new{The main} contributions of this paper are summarized as follows: 

\begin{itemize}

\item We propose a scheme called \mytitle (GDFQ), which performs 4-bitwidth quantization without any real data. \new{To our knowledge, this is the first low-bitwidth data free quantization method.}

\item \jie{We propose an effective knowledge matching generator to construct data by mining knowledge from the pre-trained full-precision model. 
The generated data retain classification boundary knowledge and data distribution.
}

\item Extensive experiments on image classification data sets demonstrate the superior performance of our method compared to the existing data-free quantization method.
\end{itemize}

\section{Related Work}
\label{Related Work}

\paragraph{\emph{\textbf{Model quantization.}}}
Model quantization targets to quantize weights, activations and even gradients to low-precision, to yield highly compact models, where the expensive multiplication operations can be replaced by additions or bit-wise operations. According to the trade-off between accuracy and computational complexity, quantization can be roughly categorized into binary neural networks (BNNs) \cite{hubara2016binarized,rastegari2016xnor,zhuang2019structured} and fixed-point quantization \cite{zhou2016dorefa,esser2019learned,zhuang2018towards}. Moreover, quantization studies mainly focus on tackling two core bottlenecks, including designing accurate quantizers to fit the categorical distribution to the original continuous distribution \cite{Cai_2017_CVPR,jung2019learning,zhang2018lq}, and approximating gradients of the non-differential quantizer during back-propagation \cite{Yang_2019_CVPR,louizos2019relaxed,zhuang2020training}.
In addition, the first practical 4-bit post-training quantization approach was introduced in \cite{banner2019post}. 
To improve the performance of neural network quantization without retraining, outlier channel splitting (OCS) \cite{zhao2019improving} proposed to move affected outliers toward the center of the distribution. 
\new{Besides,} many mainstream deep learning frameworks support 16-bit half-precision or 8-bit fixed-point quantization, such as TensorFlow \cite{abadi2016tensorflow}, PyTorch \cite{paszke2017pytorch}, PaddleSlim, etc.
In particular, these platforms provide both post-training quantization and quantization-aware training.
However, all of them require the original data.
In short, whether in scientific \new{studies} or practical applications, if the original data is unavailable, it is hard for quantization methods to work normally.

\paragraph{\emph{\textbf{Data-free model compression.}}}
Recently, the \new{researches about data-free} had included more model compression methods, such as knowledge distillation \cite{nayak2019zero,chen2019data}, low-rank approximation \cite{yoo2019knowledge} and model quantization \cite{zhao2019improving}.
A novel framework \cite{chen2019data} exploited generative adversarial networks to perform data-free knowledge distillation. Another work \cite{lopes2017data}  reconstructed a new data set based solely on the trained model and the activation statistics, and finally distilled the pre-trained ``teacher'' model into the smaller ``student'' network. 
Zero-shot knowledge distillation \cite{nayak2019zero} synthesized pseudo data by utilizing the prior information about the underlying data distribution. Specifically, the method extracted class similarities from the parameters of \new{the} teacher model and modeled the output space of the teacher network as a Dirichlet distribution. KEGNET \cite{yoo2019knowledge} proposed a novel data-free low-rank approximation approach assisted by knowledge distillation. This method contained a generator \new{that} generated artificial data points to replace missing data and a decoder \new{that} aimed to extract the low-dimensional representation of artificial data. In Adversarial Belief Matching \cite{micaelli2019zero-shot}, a generator generated data on which the student mismatched the teacher, and then the student network was trained using these data.

Quantization also faces the situation without original data while
previous quantization methods generally need original data to improve their performance. However, in some instances, it \new{is difficult} to get the original data. Recently, some \new{studies focused} on data-free model quantization.
DFQ \cite{nagel2019data} argued that data-free quantization performance could be improved by equalizing the weight ranges of each channel in the network and correcting biased quantization error.
ZeroQ \cite{Cai_2020_CVPR}, a novel zero-shot quantization framework, enabled mixed-precision quantization without any access to the training or validation data. 
However, these data-free quantization methods work well for 8-bit, but got poor performance in aggressively low bitwidth regimes such as 4-bit.

\section{Problem Definition}
\label{problem_definition}
\paragraph{\textbf{\emph{Data-free quantization problem.}}}
\jie{
Quantization usually requires original data for calibration or fine-tuning.
In many practical scenarios, the original data may not be available due to private or even confidential issues.
In this case, we cannot use any data\new{; thus,} the general scheme of the network quantization will lose efficacy and even fail to work completely, resulting in a quantized model with inferior performance.
Given a full-precision model $M$, data-free quantization aims to construct fake data $(\hat{\bx}, y)$ and meanwhile quantize a model $Q$ from the model $M_{\theta}$. 
Specifically, to compensate for the accuracy loss from quantization, training-aware quantization can fine-tune the quantized model by optimizing the following problem:
\begin{equation}
    \min_{Q, \hat{\bx}} \mmE_{\hat{\bx}, y}
    \left[\ell({Q}(\hat{\bx}), y)\right],
\end{equation}
where $\hat{\bx}$ is a generated fake sample, $y$ is the corresponding label\new{, and} $\ell(\cdot,\cdot)$ is a loss function, such as cross-entropy loss and mean squared error.
}

\paragraph{\textbf{\emph{Challenges of constructing data.}}}
Because of the absence of original data, one possible \new{way} is to construct data by exploiting information from a pre-trained full-precision model. 
Although the full-precision model may contain \new{rich} data information, such latent information alone is hard to exploit for recovering the original data.
In practice, the performance of quantization highly depends on the quality of constructed data.
With the limited information of the pre-trained model, \new{constructing} meaningful data is very challenging.

Recently, one data-free quantization method (ZeroQ) \cite{Cai_2020_CVPR} constructs fake data by using a linear operator with gradient update information.
With the help of constructed data, ZeroQ is proposed to improve the quantization performance.
However, ZeroQ has insufficient information to improve the performance of quantization with the following two issues.
First, it constructs fake data without considering label information, and thus neglects to exploit the classification boundary knowledge from the pre-trained model. 
Second, it enforces the batch normalization statistics of a single data \new{point instead of the whole data}, leading to being far away from the real data distribution.
To address these issues, one can construct a generator $G$ to produce fake data by considering label information and using \new{powerful} neural networks,
\begin{equation}
    \hat{\bx} =G(\bz|y),\quad \bz \sim p(\bz), 
\end{equation}
where $\bz$ is a random vector drawn from some prior distribution $p(\bz)$, \eg Gaussian distribution or uniform distribution.
By using the generator, we are able to construct fake data to improve quantization performance. 
However, what knowledge in the pre-trained model can be exploited and how to learn a good generator remain to be answered.

\begin{figure*}[t]
\vskip 0.2in
\begin{center}
\centerline{\includegraphics[width=1\linewidth]{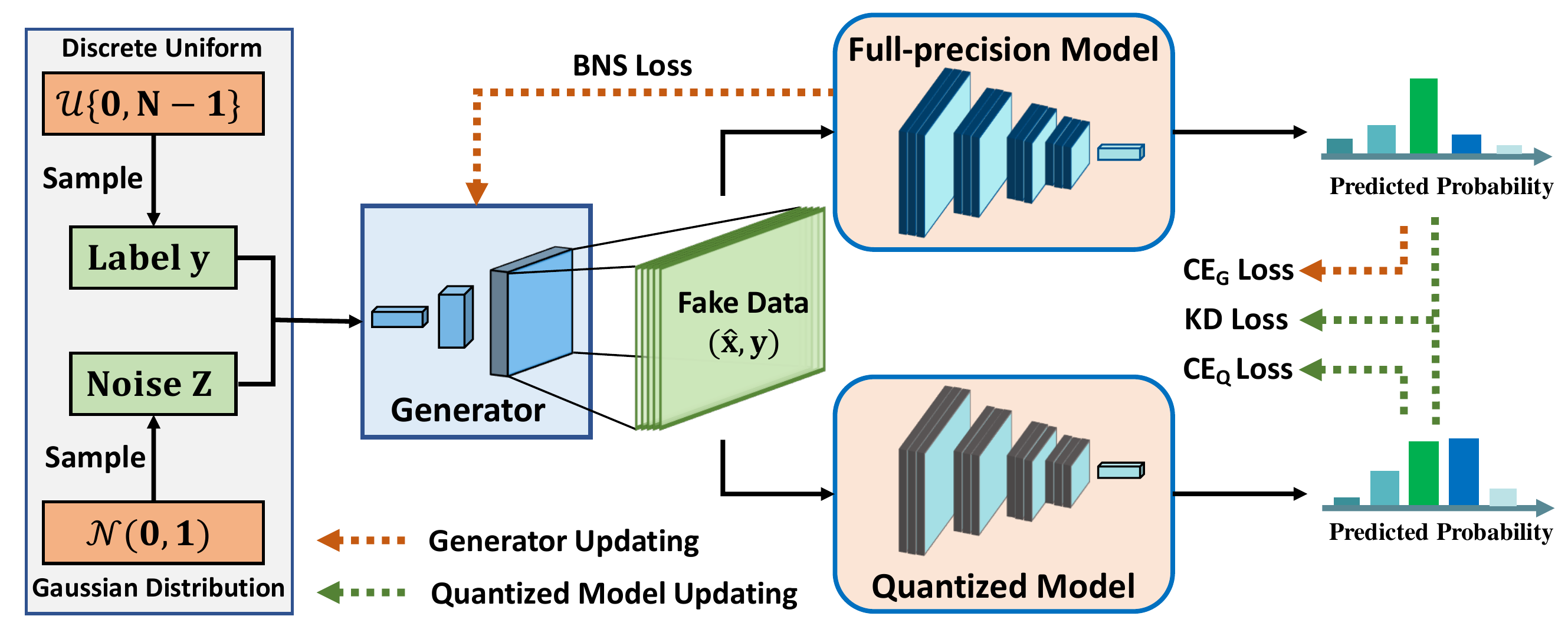}}
\caption{An overview of the proposed method. Given Gaussian noise and \new{the} label as input, the generator creates fake data and feeds them into both \new{the} full-precision model and \new{the} quantized model. The fixed full-precision model provides knowledge for updating the generator. The quantized model learns latent knowledge from the generator and the full-precision model.}
\label{framework}
\end{center}
\vskip -0.2in
\end{figure*}

\section{{\mytitle}}
\label{Method}

\jie{
In many practical scenarios, training data are unavailable\new{; thus,}  an existing method \cite{Cai_2020_CVPR} performs data-free quantization by constructing data.
\wrong{However, without exploiting knowledge from a pre-trained model, directly \new{constructing} data lacks label information and leads to \new{being} far away from the real data distribution.}
To address \new{this issue}, we aim to design a generator to produce fake data.
Then, we are able to perform supervised learning to improve the performance of quantization. 
The overall framework is shown in Fig. \ref{framework}.
}

\subsection{\jie{Knowledge Matching Generator}}
\label{generator}
\jie{
When training a deep neural network, it captures sufficient data information to make a decision \cite{zhang2016understanding}. 
In this sense, the pre-trained DNN contains some knowledge information of the training data, \eg classification boundary information and distribution information.
We therefore elaborate how to effectively construct informative data from the pre-trained model.
}

\paragraph{\emph{\textbf{\jie{Classification boundary information matching.}}}}
\jie{
The pre-trained model contains classification boundary information.
Unfortunately, such information is \new{difficult} to \new{exploit} to recover the data \new{near to} the classification boundary. 
Recently, generative adversarial networks (GANs) \cite{goodfellow2014generative,cao2018adversarial,cao2019multi} have achieved considerable success \new{in producing} data.
\new{Since the real data are unavailable, the discriminator in a GAN cannot work in the data-free quantization task.}
Without the discriminator, \new{learning a generator to produce meaningful data is difficult.}
}

\jie{In this paper, we propose a generator to produce fake data for \new{the} data-free quantization task.}
For this task, although the original data cannot be observed, we are able to easily confirm the number of categories of original data by the last layer in a pre-trained model.
\jie{As shown in Fig. \ref{motivation} (a),} different categories of data should be distributed in different data spaces.
To generate fake data, we introduce a noise vector $\bz$ conditioned on a label $y$.
Here, we sample noise from Gaussian distribution $\mN({0},{1})$ and uniformly sample a label from $\{0, 1, ..., n-1\}$.
\jie{Then, the generator maps a prior input noise vector $\bz$ and the given label $y$ to \new{the} fake data $\hat{\bx}$.
Formally, we define the knowledge matching generator as follows:}
\begin{eqnarray}\label{eq:generate_fake_image}
\hat{\bx}&=&G(\bz|y),\quad \bz \sim \mN({0},{1}).
\end{eqnarray}
To improve the quantization performance, the generator should have the ability to generate data that are effective to fine-tune a quantized model.
\jie{To this \new{end}, given the Gaussian noise $\bz$ and the corresponding label $y$, the generated fake data should be classified to the same label $y$ by the full-precision pre-trained model $M$.}
Therefore, we introduce the following cross-entropy loss $\CE(\cdot,\cdot)$ to train the generator $G$:
\begin{eqnarray}\label{eq:generate_onehot_loss}
\mL^G_\onehot(G)=\mmE_{\bz, y}\left[\CE(M(G(\bz|y)), y)\right].
\end{eqnarray}

\paragraph{\emph{\textbf{\jie{Distribution information matching.}}}}
\jie{
\new{In addition}, the pre-trained model contains the distribution information of training data.
Such distribution information can be captured by batch normalization \cite{ioffe2015batch}, which is used to control the change of the distribution.
\xsk{While training the full-precision model, the batch normalization statistics (BNS), \ie the mean and variance, are computed dynamically.
For every batch, batch normalization (BN) layers only compute statistics on the current mini-batch inputs and accumulate them with  momentum.
\new{Finally}, the exponential moving average (EMA) batch normalization statistics will be obtained and then they will be used in network validation and test.}
}

\jie{
To retain the BNS information, the mean and variance of the generated distribution should be the same as that of the real data distribution.}
\jie{
To this end, we learn a generator $G$ with the BNS information ($\bmu_l$ and $\bsigma_l$) encoded in the $l$-th BN layer of the pre-trained model using $\mL_\bns$:
}
\begin{eqnarray}
\mL_\bns(G)=\sum_{l=1}^L\|\bmu_l^r-\bmu_l\|_2^2+\|\bsigma_l^r-\bsigma_l\|_2^2,
\end{eqnarray}
where $\bmu_l^r$ and $\bsigma_l^r$ are the mean and variance of the fake data’s distribution at the $l$-th BN layer, \new{respectively,} and $\bmu_l$ and $\bsigma_l$ are the corresponding mean and variance parameters stored in the $l$-th BN layer of the pre-trained full-precision model\new{, respectively}.
In this way, we are able to \new{learn} a good generator to keep the distribution information from the training data.

\subsection{Fake-data Driven Low-bitwidth Quantization}
\label{finetune}

\jie{With the help of the generator, 
the data-free quantization problem can be turned to a supervised quantization problem.
Thus, we are able to quantize a model using produced meaningful data. 
However, \new{transferring} knowledge from a pre-trained model to a quantized model is difficult.
To address this, we introduce a fake-data driven quantization method and solve the optimization problem by \new{exploiting} knowledge from the pre-trained model. 
}

\paragraph{\textbf{\emph{\jie{Quantization.}}}}

Network quantization maps full-precision (32-bit) weights and activations to low-precision ones, \eg 8-bit fixed-point integer. 
We use a simple-yet-effective quantization method refer to \cite{jacob2018quantization} for both weights and activations.
\xsk{Specifically, given full-precision weights ${\theta}$ and the quantization precision $k$, we quantize ${\theta}$ to ${\theta_q}$ in the symmetric $k$-bit range:}
\begin{equation}
\label{eq:clip}
\theta_{q} = 
\begin{cases}
    -2^{k-1},  &\text{if $\theta' {<} -2^{k-1}$ }\\
    2^{k-1}{-}1, &\text{if $\theta' {>} 2^{k-1}{-}1$ }\\
    \theta', &\text{otherwise},
    \end{cases}
\end{equation}
where $\theta'$ are discrete values mapped by a linear quantization, \ie $\theta' =\left\lfloor\Delta \cdot {\theta} - b\right\rceil$,
$\left\lfloor \cdot \right\rceil$ is the round function,
and $\Delta$ and $b$ can be computed by
$\Delta=\frac{2^{k}-1}{u-l} $ and $ b = l \cdot \Delta + 2^{k-1}$.
Here, $l$ and $u$ can be set as the minimum and maximum of the floating-point weights $\theta$, respectively.

\paragraph{\emph{\textbf{\wrong{Optimization problem.}}}}
\jie{
When the real training data are unavailable, quantization may suffer from some limitations.
First, \new{direct quantization} from a full-precision model may result in severe performance degradation.
To address this issue, we aim to train the quantized model to approximate the full-precision model through the fine-tuning process.
To this end, a well fine-tuned quantized model $Q$ and the pre-trained model $M$ should classify the fake data correctly.
}
For this purpose, we use the cross-entropy loss function $\CE(\cdot,\cdot)$ to update $Q$:
\begin{eqnarray}\label{eq:quantize_onehot_loss}
\mL^Q_\onehot(Q)=\mmE_{\hat{\bx}, y}\left[\CE(Q({\hat{\bx}}), y)\right].
\end{eqnarray}
By minimizing Eq. (\ref{eq:quantize_onehot_loss}), the quantization model can be trained with the generated data to perform multi-class classification.

Second, the traditional fine-tuning process with a common classification loss function is insufficient because the data are fake.
\jie{\new{However,} with the help of fake data, we are able to apply knowledge distillation \cite{hinton2015distilling} to improve the quantization performance.
Specifically, 
given the same inputs, the outputs of a quantized model and full-precision model should be close enough to guarantee that the quantized model is able to achieve nearly performance compared with the full-precision model.
Therefore, we utilize knowledge distillation to recover the performance of the quantized model by using fake data $\hat{\bx}$ to simulate the training data.
}
Then, the quantized model can be fine-tuned using the following Kullback-Leibler loss \new{function} $\KL(\cdot,\cdot)$:
\xsk{\begin{eqnarray}\label{eq:Q_loss}
\mL_\kd(Q)=\mmE_{\hat{\bx}}
    \left[\KL(Q({\hat{\bx}}), M({\hat{\bx}}))\right].
\end{eqnarray}}%
By optimizing \new{the loss in} (\ref{eq:Q_loss}), the quantization model can learn knowledge from the full-precision model.

\paragraph{\emph{\textbf{\xsk{Fine-tuning with fixed BNS.}}}}

\jie{
To stabilize the training, we fix the batch normalization statistics (BNS) in the quantized model during fine-tuning. In this way, the BNS in the quantized model \new{are} corresponding to the statistics of the real data distribution.
With the help of fixed BNS, the quantized model always maintains the real data information to improve quantization performance. 
}

\begin{algorithm}[t]
\caption{\mytitle}
\label{alg:alg1}
\textbf{Input}: Pre-trained full-precision model $M$.\\
\textbf{Output}: Generator $G$, quantized model $Q$.

 \begin{algorithmic}[l]
    \STATE \wrong{Update $G$ several times as a warm-up process.}
    \STATE Quantize model $M$ using Eq. (\ref{eq:clip}), get quantized model $Q$.
    \STATE Fix the batch normalization statistics in all BN layers of quantized model $Q$.
    \FOR {$ t= 1, \ldots, T_{fine-tune} $}
    \STATE Obtain random noise $\bz \sim \mN (0,1)$ and label $y$.
	\STATE Generate fake image $\hat{\bx}$ using Eq. (\ref{eq:generate_fake_image}).
	\STATE Update generator $G$ by minimizing Loss (\ref{eq:loss_G}).
	\STATE Update quantized model $Q$ by minimizing Loss (\ref{eq:loss_D}).
    \ENDFOR
	\end{algorithmic}
	\label{alg:AT}
\end{algorithm}

\subsection{Training Process}
\label{train_process}
\jie{We propose a low-bitwidth quantization algorithm to alternately optimize the generator $G$ and the quantized model $Q$.
In our alternating training strategy, the generator is able to generate different data with each update.
By increasing the diversity of data, the quantized model $Q$ can be trained to improve the performance.
\new{In addition}, to make the fine-tuning of $Q$ more stable, we firstly train $G$ solely several times as a warm-up process.
The overall process is shown in Algorithm~\ref{alg:alg1}.
In contrast, one can train the generator $G$ with a full-precision model, and then fix the generator to train the quantized model $Q$ until convergence.
In this separate training strategy, when the diversity of the generated data is poor, the quantized model has a limitation to improve the quantization performance. 
}

\paragraph{\emph{\textbf{\jie{Training generator $G$.}}}}
First, we randomly sample a Gaussian noise vector $\bz \sim \mN (0,1)$ with a label $y$. 
Then we use $G$ to generate fake data from the distribution and update $G$. 
The final generator loss $\mL_{1}(G)$ is formulated as follows:
\begin{eqnarray}
\label{eq:loss_G}
\mL_{1}(G)&=&\mL^G_\onehot(G) + \beta\mL_\bns(G),
\end{eqnarray}
where $\beta$ is a trade-off parameter.

\paragraph{\emph{\textbf{\jie{Training quantized model $Q$.}}}}
We quantize the model according to Eq. (\ref{eq:clip}).
Then, we replace the BNS in the quantized model \new{with} the fixed batch normalization statistics (FBNS) as described in Section~\ref{finetune}.
So far, the quantized model has inherited the information contained in BNS and a part of latent knowledge from the parameters of the pre-trained model.
In the fine-tuning process, we train the $G$ and $Q$ alternately in every epoch.
Based on the warmed up $G$, we obtain fake samples and use them to optimize the quantized model $Q$. The final quantized model loss function $\mL_{2}(Q)$ is formulated as follows:
\begin{eqnarray}
\label{eq:loss_D}
\mL_{2}(Q)=\mL^Q_\onehot(Q)+\gamma\mL_{\kd}(Q),
\end{eqnarray}
where $\gamma$ is a trade-off parameter.
We do not stop updating $G$ because if we have a better $G$,  the fake data will be \new{more similar to} real training data and the upper limit of optimizing $Q$ will be improved.
Note that we keep the pre-trained full-precision model fixed \new{at all times}.

\section{Experiments}
\setcounter{footnote}{0}

\subsection{Data Sets and Implementation Details}
\jing{
We evaluate the proposed method on well-known data sets including CIFAR-10~\cite{krizhevsky2009learning}, CIFAR-100~\cite{krizhevsky2009learning}, and ImageNet~\cite{deng2009imagenet}.} CIFAR-10 consists of 60k images from 10 categories, with 6k images per category. There are 50k images for training and 10k images for testing. CIFAR-100 has 100 classes and each class contains 500 training images and 100 testing images. ImageNet is one of the most challenging and largest benchmark data sets for image classification, which has around 1.2 million real-world images for training and 50k images for validation. 

\jing{
Based on the full-precision pre-trained models from pytorchcv\footnote{https://pypi.org/project/pytorchcv/}, we quantize ResNet-20~\cite{he2016deep} on CIFAR-10/100 and ResNet-18~\cite{he2016deep}, BN-VGG-16~\cite{SimonyanZ14a} and Inception v3~\cite{szegedy2016rethinking} on ImageNet.} \new{In all experiments,} we quantize all layers including the first and last layers of the network following~\cite{Cai_2020_CVPR} and \new{the activation clipping values are per-layer granularity.} \lhk{All implementations are based on PyTorch.}

For CIFAR-10/100, we construct the generator following ACGAN \cite{odena2017conditional} and the dimension of noise is 100. 
During training, we optimize the generator and quantized model using  Adam~\cite{KingmaB14} and SGD with Nesterov~\cite{nesterov1983method} respectively, where the momentum term and weight decay in Nesterov are set to $0.9$ and $1 \times 10^{-4}$.
Moreover, the learning rates of quantized models and generators are initialized to $1 \times 10^{-4}$ and $1 \times 10^{-3}$, respectively. 
Both of them \xsk{are decayed by 0.1 for every 100 epochs.}
\new{In addition}, we train the generator and quantized model for 400 epochs with 200 iterations per epoch. 
\lhk{To obtain a more stable clip range for activation, we calculate the moving average of activation's range in the first four epochs without updating the quantized models and then fix this range for subsequent training.}
For $\mL_{1}(G)$ and $\mL_{2}(Q)$, we set $\beta=0.1$ and $\gamma=1$ after a simple grid search.
For ImageNet, we replace the generator's standard batch normalization layer with the categorical conditional batch normalization layer for fusing label information following SN-GAN \cite{miyato2018spectral} and set the initial learning rate of the quantized model as $1 \times 10^{-6}$. Other training settings are the same as those on CIFAR-10/100.

\subsection{Toy Demonstration for Classification Boundary Matching}
To evaluate that the fake data generated by our $G$ are able to match the classification boundary information, we design a toy experiment. The results are shown in Fig. \ref{motivation}.
First, we create a toy binary classification data set by uniform sampling from $-4$ to $+4$, and the label is shown in Fig. \ref{motivation} (a). 
Second, we construct a simple neural network $T$, which is composed of several linear layers, BN layers, and ReLU layers, and we train it using the toy data. The classification boundaries are shown in each subfigure.
To simulate the process of our method, we sample noises from Gaussian distribution and every noise has a random label \new{of} 0 or 1 (Fig. \ref{motivation} (b)).
Then, we generate fake data from noises by learning from the pre-trained model $T$.
Fig. \ref{motivation} (c) and Fig. \ref{motivation} (d) show the fake data generated by the ZeroQ method and our method,  respectively. 
The data generated by ZeroQ do not capture the real data distribution since it neglects the inter-class information; while our method is able to produce fake data that not only have label information but also match the classification boundaries.

\subsection{Comparison \new{of the} Results}
To further evaluate the effectiveness of our method, we include the following methods for study. \lhk{\textbf{FP32:} the full-precision pre-trained model. \textbf{FT:} we use real training data instead of fake data to  fine-tune the quantized model by minimizing $\mL_{2}$. \textbf{ZeroQ:} a data-free post-training quantization method. We obtain the result from the publicly released code of ZeroQ~\cite{Cai_2020_CVPR}.}

We quantize both weights and activations to 4-bit and report the comparison results in Table \ref{tb:comparison_CIFAR_ImageNet}. For CIFAR-10, our method achieves much higher accuracy than \new{that of} ZeroQ~\cite{Cai_2020_CVPR}. When the number of categories increases in CIFAR-100, our method suffers a much smaller degradation in accuracy compared with \new{that of} ZeroQ. \lhk{The main reason is that, our method gains more prior knowledge from the full-precision model.} These results demonstrate the effectiveness of our method on simple data sets with 4-bit quantization. 
\lhk{\new{For} large scale and categories data set\new{, such as} ImageNet, existing data-free quantization methods suffer from severe performance degradation. However, our generated images contain abundant category information and similar distribution with real data. As a result, our method recovers the accuracy of quantized models significantly with the help of generated fake data and knowledge distillation on three typical networks.}
\new{More experiments on different models and methods can be found in the supplementary material.}

\begin{table}[t]
\renewcommand\arraystretch{1.1}
\renewcommand{\tabcolsep}{5.0pt}
\begin{center}
\caption{Comparisons on CIFAR-10/100 and ImageNet. We report the average and standard deviation of our method to show that our method is stable. We quantize both \new{the} weights and activations of \new{the} models to 4-bits and report the top1 accuracy.}
\label{tb:comparison_CIFAR_ImageNet}
\begin{tabular}{c|c|cc|cc}
\hline
\multirow{2}{*}{Data Set} &\multirow{2}{*}{Model} & \multicolumn{2}{c}{Real Data} &\multicolumn{2}{|c}{Data Free }  \\ \cline{3-6}
 &  & FP32 & FT & ZeroQ~\cite{Cai_2020_CVPR} & Ours  \\ 
\hline
CIFAR-10  & ResNet-20 & 94.03 & 93.11 & 79.30 & \textbf{90.25 $\pm$ 0.30} \\
CIFAR-100 & ResNet-20 & 70.33 & 68.34 & 45.20 & \textbf{63.58 $\pm$ 0.23} \\
\hline
\multirow{3}{*}{ImageNet} & BN-VGG16 & 74.28 & 68.83 & 1.15 & \textbf{67.10 $\pm$ 0.29} \\
& ResNet-18 & 71.47 & 67.84 & 26.04 & \textbf{60.60 $\pm$ 0.15} \\
& Inception v3 & 78.80 & 73.80 & 26.84 & \textbf{70.39 $\pm$ 0.20}\\
\hline
\end{tabular}
\end{center}
\end{table}

\subsection{Ablation Studies}
\lhk{In this section, we first evaluate the effectiveness of each component in $\mL_{1}$ and  $\mL_{2}$. Second, we explore how fixed BNS affects our method. Then, we compare our method with different quantization methods. Last, we further study the effect of different stopping conditions. All the ablation experiments are conducted on the CIFAR-10/100 data sets.}

\begin{table}[t]
\tabcolsep=10pt
\begin{center}
\caption{\jing{Effect of different loss functions} of generator $G$. We \jing{quantize both} \new{the} weights and activations of \new{the} models to 4-bits and report the top1 accuracy on CIFAR-100.}
\label{tb:ablation_lossG}
\begin{tabular}{c|cc|c}
\hline
Model &  CE Loss  &  BNS Loss & Acc. ($\%$)\\
\hline
\multirow{4}{*}{\begin{tabular}[c]{@{}c@{}}ResNet-20\\ (4-bit)\end{tabular}} &   $\bf\times$    &   $\bf\times$    &  30.70 \\ 

      &\checkmark   &   $\bf\times$  &  54.51 \\
      &$\bf\times$   &  \checkmark  &  44.40 \\

      & \checkmark      & \checkmark  & \textbf{63.91} \\ 
\hline
\end{tabular}
\end{center}
\end{table}

\begin{table}[t]
\tabcolsep=10pt
\begin{center}
\caption{\jing{Effect of different} loss functions of $Q$. We keep the weights and activations of \new{the} models to 4-bits and report the top1 accuracy on CIFAR-100.}
\label{tb:ablation_lossQ}
\begin{tabular}{c|cc|c}
\hline
Model & CE Loss& KD Loss  & Acc. ($\%$)\\
\hline
\multirow{3}{*}{\begin{tabular}[c]{@{}c@{}}ResNet-20\\ (4-bit)\end{tabular}}  &\checkmark  &   $\bf\times$  & 55.55 \\ 
  &        $\bf\times$         & \checkmark  &  62.98 \\ 
  &        \checkmark         & \checkmark  & \textbf{63.91} \\ 

\hline
\end{tabular}
\end{center}
\end{table}

\paragraph{\emph{\textbf{Effect of different losses.}}}
To verify the effectiveness of different components in our method, we conduct a series of ablation experiments on CIFAR-100 with ResNet-20. 
\lhk{Table \ref{tb:ablation_lossG} reports the top-1 accuracy of quantized models with different components of $\mL_{1}$. In this ablation experiment, we fine-tune quantized models with complete $\mL_{2}$. Since we do not use both CE loss and BNS loss, we have no way to optimize $G$, which means we use the fake data generated from the initial $G$ to fine-tune the quantized model. In this case, the distribution of the fake data is far away from that of original data because the generator \new{receives} no guidance from the full-precision model. Therefore, the quantized model suffers \jing{from a large} performance degradation. To utilize the knowledge in the full-precision model, we use CE loss to optimize $G$ and achieve a better quantized model. In this case, the generator produces fake data that can be classified with high confidence by the full-precision model. Last, we combine CE loss and BNS loss with a coefficient and achieve the best result. The BNS loss encourages the generator to generate fake data that match the statistics encoded in full-precision model's BN layers so that these fake data have \new{a} much similar distribution with real data. In summary, \jing{both CE loss and BNS loss contribute to better performance of the quantized model.}}

We further conduct ablation experiments to analyze the effectiveness of each component in $\mL_{2}$.
\lhk{Table \ref{tb:ablation_lossQ} reports the top-1 accuracy of quantized models with different components of $\mL_{2}$. In this experiment, we optimize the generator with complete $\mL_{1}$. When only introducing KD loss, the quantized model receives knowledge from the full-precision model's prediction and achieves 62.98$\%$ on top-1 accuracy. To use the additional label information, we combine BNS loss with CE loss. \jing{The resultant model achieves a 0.93\% improvement on top-1 accuracy. }}

\paragraph{\emph{\textbf{Effect of the fixed BNS.}}}
To verify the effectiveness of fixing batch normalization, we conduct ablation studies with ResNet-20 on CIFAR-10/100. The results are shown in Table \ref{tb:ablation_BN}. 
\lhk{When we fix batch normalization statistics during fine-tuning, we narrow the statistics gap between the quantized model and the full-precision model. As a result, we achieve a much higher top-1 accuracy than that with standard batch normalization. }

\begin{table}[t]
\tabcolsep=20pt
\begin{center}
\caption{Ablation experiments on the fixed BNS (FBNS). We keep the weights and activations of \new{the} models to be 4-bits and report the top1 accuracy on CIFAR-10/100. We use ``w/o FBNS'' to represent that we use fake data to fine-tune the quantized models without fixed BNS. Similarly, we use ``w/ FBNS'' to represent the fine-tuning process with fixed BNS.}
\label{tb:ablation_BN}
\begin{tabular}{ccc}
\hline
Data Set & w/o FBNS & w/ FBNS \\ \hline 
CIFAR-10          & 89.21         & \textbf{90.23}    \\ 
CIFAR-100         & 61.12         & \textbf{63.91}    \\ \hline
\end{tabular}
\end{center}
\end{table}

\begin{table}[t]
\tabcolsep=8pt
\begin{center}
\caption{Comparison  of  different  post-training  quantization  methods. We use real data sets as calibration sets and report the accuracy of the quantized model without fine-tuning.
}
\label{tb:post_training_experiments}
\begin{tabular}{ccccccc}
\hline
Data Set & Model & Method  & W8A8  & W6A6  & W5A5  & W4A4  \\ \hline
\multirow{4}{*}{CIFAR-10} & \multirow{4}{*}{\begin{tabular}[c]{@{}c@{}}ResNet-20\\ (94.03)\end{tabular}} & MSE \cite{sung2015resiliency} & 93.86 & 93.10  & 91.38 & 81.59 \\
 &  & ACIQ \cite{banner2018aciq}        & 93.69 & 92.22 & 86.66 & 61.87 \\
 &  & KL \cite{chen2015mxnet}          & 93.72 & 92.32 & 90.71 & 80.05 \\
 &  & Ours & \textbf{93.92} & \textbf{93.38} & \textbf{92.39} & \textbf{85.20}  \\ \hline
\multirow{4}{*}{CIFAR-100} & \multirow{4}{*}{\begin{tabular}[c]{@{}c@{}}ResNet-20\\ (70.33)\end{tabular}} & MSE \cite{sung2015resiliency} & 70.11 & 66.87 & 60.49 & 27.11 \\
 &  & ACIQ \cite{banner2018aciq}        & 69.29 & 63.21 & 48.21 & 8.72  \\
 &  & KL \cite{chen2015mxnet}          & 70.15 & 67.65 & 57.55 & 15.83 \\
 &  & Ours & \textbf{70.29} & \textbf{68.63} & \textbf{64.03} & \textbf{43.12} \\ \hline
\multirow{4}{*}{ImageNet}  & \multirow{4}{*}{\begin{tabular}[c]{@{}c@{}}ResNet-18\\ (71.47)\end{tabular}} & MSE \cite{sung2015resiliency} & 71.01 & 66.96 & 54.23 & 15.08 \\
 &  & ACIQ \cite{banner2018aciq}        & 68.78 & 61.15 & 46.25 & 7.19  \\
 &  & KL \cite{chen2015mxnet}          & 70.69 & 61.34 & 56.13 & 16.27 \\
 &  & Ours & \textbf{71.43} & \textbf{70.43} & \textbf{64.68} & \textbf{33.25} \\ \hline
\end{tabular}
\end{center}
\end{table}

\subsection{Further Experiments}
\paragraph{\emph{\textbf{Comparisons with different post-training quantization methods.}}}
We compare different post-training quantization methods \cite{sung2015resiliency,banner2018aciq,chen2015mxnet} on CIFAR-10/100 and ImageNet and \jing{show the results} in Table \ref{tb:post_training_experiments}. In this experiment, we use images from real data sets as calibration sets \new{for} quantized models with different post-training quantization methods. 
Specifically, we compare our quantization method with MSE(mean squared error), ACIQ, and KL(Kullback-Leibler), which are popular methods to decide the clip values of weight and activation. 
Our method shows much better performance than \new{that of the} other methods, which means it is more suitable in low-bitwidth quantization. 
\new{Furthermore}, the experimental results show that as the data set gets larger, the accuracy decreases. 
With the decrease of precision, all \new{the} quantization methods behave \new{more poorly}. \new{Specifically,} when the precision drops to 4-bit, the accuracy declines sharply.

\begin{table}[t]
\tabcolsep=15pt
\begin{center}
\caption{Comparison of separate training and alternating training of $G$ and $Q$.}
\label{tb:alternately_training}
\begin{tabular}{cc}
\hline
Training Strategy & Acc. ($\%$) \\ \hline
Separate Training & 61.81 \\
Alternating Training        & \textbf{63.91}             \\
\hline
\end{tabular}
\end{center}
\end{table}

\paragraph{\emph{\textbf{Effect of two training strategies.}}}
We investigate the effect of two kinds of training strategies.
1) Training generator and quantized model \jing{in two steps. We first train the generator by minimizing \new{the loss} (\ref{eq:loss_G}) until convergence}. Then, we train the quantized model by minimizing \new{the loss} (\ref{eq:loss_D}).
2) Training the generator and quantized model alternately in each iteration following Algorithm \ref{alg:AT}.
From the results in Table \ref{tb:alternately_training}, alternating training performs significantly better than separate training.
\new{Therefore,} we use alternating training in other experiments.

\begin{table}[t]
\tabcolsep=8pt
\begin{center}
\caption{Effect of different thresholds in the stopping condition of $G$.}
\label{tb:early_stop_G}
\begin{tabular}{cccccc}
\hline
Threshold $\eta$ ($\%$) & 90.00 & 95.00 & 99.00 & 99.50 & w/o stopping condition \\ 
\hline
Acc. ($\%)$ & 57.20 & 57.56 & 59.09 & 59.67 & \textbf{63.91} \\
\hline
\end{tabular}
\end{center}
\end{table}
\paragraph{\emph{\textbf{Effect of different thresholds in stopping conditions.}}}

\lhk{In this experiment, we stop the training of the generator if the classification accuracy of the full-precision model on fake data is larger than a threshold $\eta$. Table \ref{tb:early_stop_G} reports the results of different thresholds $\eta$ in stopping condition. When increasing the threshold, the generator is trained with quantized models for more epochs, and we get a better fine-tuning result. \new{We} achieve the best performance when we do not stop optimizing the generator. These results demonstrate that optimizing the generator and quantized model simultaneously increases the diversity of data, which is helpful for fine-tune quantized models.}

\section{Conclusion}
In this paper, we have proposed a \mytitle scheme to eliminate the data dependence of quantization methods. 
First, we have constructed a knowledge matching generator to produce fake data for the fine-tuning process. 
The generator is able to learn the classification boundary knowledge and distribution information from the pre-trained full-precision model.
\new{Next,} we have quantized the full-precision model and fine-tuned the quantized model using the fake data.
Extensive experiments on various image classification data sets have demonstrated the effectiveness of our data-free method.

\section*{Acknowledgements}
This work was partially supported by the Key-Area Research and Development Program of Guangdong Province 2018B010107001, Program for Guangdong Introducing Innovative and Entrepreneurial Teams 2017ZT07X183, Fundamental Research Funds for the Central Universities D2191240.

\clearpage
\appendix

\begin{center}
\Large{\textbf{Generative Low-bitwidth Data Free Quantization} \\ (Supplementary Materials)}
\end{center}

\section{More Experimental Results}
\subsection{Experiments on More Networks}

We conduct more experiments on different networks and report the results in Table \ref{tb:comparison_morenet}. From the table, our method achieves much better performance than ZeroQ \cite{Cai_2020_CVPR} in terms of accuracy at different precisions. More importantly, our method shows larger performance gain over ZeroQ at lower bit-width (e.g., W4A4) for all models. These results demonstrate the effectiveness of our method.

\begin{table}[h]
\renewcommand\arraystretch{1.1}
\renewcommand{\tabcolsep}{5.0pt}
\begin{center}
\caption{Experiments on more networks on ImageNet. We quantize ResNet-50, MobileNetV2, and ShuffleNet to 6-bit and 4-bit, and report the Top-1 accuracy.}
\label{tb:comparison_morenet}
\begin{tabular}{c|ccc|ccc|ccc}
\hline
      & \multicolumn{3}{c|}{ResNet-50}          & \multicolumn{3}{c|}{MobileNet V2}       & \multicolumn{3}{c}{ShuffleNet}         \\ \cline{2-10} 
      & FP32  & W6A6           & W4A4           & FP32  & W6A6           & W4A4           & FP32  & W6A6           & W4A4           \\ \hline
ZeroQ & 77.72 & 72.86          & 0.12           & 73.03 & 69.62          & 3.31           & 65.07 & 46.25          & 0.27           \\
Ours  & 77.72 & \textbf{76.59} & \textbf{55.65} & 73.03 & \textbf{70.98} & \textbf{51.30} & 65.07 & \textbf{60.12} & \textbf{21.78} \\ \hline
\end{tabular}
\end{center}
\end{table}

\subsection{Comparisons with DFQ}

We implement the algorithm in DFQ \cite{nagel2019data} and report the results in Table \ref{tb:comparison_dfq}. Our method achieves much higher performance than DFQ, especially in low bit-width.

\begin{table}[h]
\renewcommand\arraystretch{1.1}
\renewcommand{\tabcolsep}{5.0pt}
\begin{center}
\caption{Comparisons with the DFQ method. We quantize MobileNetV2 and ResNet-18 to 8-bit, 6-bit and 4-bit, and report the Top-1 accuracy on ImageNet.}
\label{tb:comparison_dfq}
\begin{tabular}{c|cccc|cccc}
\hline
     & \multicolumn{4}{c|}{MobileNet V2}               & \multicolumn{4}{c}{ResNet-18}          \\ \cline{2-9} 
     & FP32  & W8A8  & W6A6           & W4A4           & FP32  & W8A8  & W6A6  & W4A4           \\ \hline
DFQ  & 73.03 & 66.06 & 52.82          & 0.11           & 71.74 & 70.13 & 14.67 & 0.10           \\
Ours & 73.03 & 72.80 & \textbf{70.98} & \textbf{51.30} & 71.74 & 70.68 & 70.13 & \textbf{60.52} \\ \hline
\end{tabular}
\end{center}
\end{table}

\nocite{langley00}

\bibliographystyle{splncs04}
\bibliography{egbib}

\end{document}